%% file: main.tex
\documentclass[10pt,journal,final,a4paper]{IEEEtran}
\def\eg{\textit{e.g.}}
\def\ie{\textit{i.e.}}

\usepackage{array}
\usepackage[caption=false,font=normalsize,labelfont=sf,textfont=sf]{subfig}
\usepackage{url}
\usepackage{verbatim}
\usepackage{graphicx}
\usepackage{cite}
\usepackage{lipsum}
\usepackage{balance}

\usepackage{tikz,xcolor}
 
\usepackage{xspace}
\usepackage{caption}
\usepackage{amsmath,amsfonts}
\usepackage{array}
\usepackage{multirow}
\usepackage{bbding}
\usepackage{array}
\usepackage{makecell}
\usepackage{amsmath}
\usepackage{xspace}
\usepackage{color}
\usepackage{amssymb}
\usepackage{booktabs}

\usepackage{hyperref}
 
\def\name{\emph{LSEF}\xspace}
\def\MA{\emph{SEM}\xspace}
\def\MB{\emph{DDM}\xspace}
\def\MC{\emph{CIM}\xspace}
\def\MD{\emph{RAO}\xspace}

\usepackage{CJKutf8}

\usepackage{array}
 
\usepackage[ruled,vlined]{algorithm2e}
\usepackage{multirow}
\usepackage{bbding}
 
\usepackage{makecell}
\usepackage{amsmath}
\usepackage{xspace}
\usepackage{color}

\usepackage{amssymb}
\usepackage{booktabs}
\usepackage{pifont}
\usepackage{bm}
\usepackage{soul}
\usepackage{hyperref}
\hypersetup{
    colorlinks=true, 
    urlcolor=magenta,   
}

\begin{document}

\title{Robust Low-Rank Sparse Framework for Video-Based Affective Computing}

\author{Feng-Qi Cui, Jinyang Huang\textsuperscript{$\ast$}, Sirui Zhao, Xinyu Li, Xin Yan, Ziyu Jia, Xiaokang Zhou
\thanks{Feng-Qi Cui and Sirui Zhao are with the State Key Laboratory of Cognitive Intelligence, School of Computer Science and Technology, University of Science and Technology of China, Hefei, China.} 
\thanks{Feng-Qi Cui, Jinyang Huang, Xinyu Li are with the Anhui Province Key Laboratory of Affective Computing and Advanced Intelligent Machine, and the School of Computer Science and Information Engineering, Hefei University of Technology, Hefei, China.}
\thanks{Xin Yan is with Cylingo Group, Beijing, China.}
\thanks{Ziyu Jia is with the Beijing Key Laboratory of Brainnetome and Brain-Computer Interface, and the Brainnetome Center, Institute of Automation, Chinese Academy of Sciences, Beijing, China.}
\thanks{Xiaokang Zhou is with the Faculty of Business Data Science, Kansai University, Osaka 565-8585, Japan, and also with the RIKEN Center for Advanced Intelligence Project, RIKEN, Tokyo 103-0027, Japan}

\thanks{Corresponding author\textsuperscript{$\ast$}: Jinyang Huang (Email: hjy@hfut.edu.cn).}

}

\markboth{Journal of \LaTeX\ Class Files,~Vol.~18, No.~9, September~2020}%
{How to Use the IEEEtran \LaTeX \ Templates}

\maketitle

\begin{abstract}
Video-based Affective Computing (VAC) is central to emotion understanding, human--computer interaction, and emerging consumer electronics applications, yet it often exhibits unstable optimization and representational degradation when facing complex, non-stationary affective dynamics in the wild.
A key challenge is that the same short-term behavioral fluctuation can convey different affective meanings under different long-term emotional contexts; however, most existing VAC models implicitly entangle heterogeneous affective factors into a single latent space, leading to component mixing, coupled propagation, and cross-scale inconsistency.
We argue that effective VAC requires a hierarchical structural mechanism that explicitly disentangles distinct affective components, \ie, emotional bases describing the long-term emotional tone and transient fluctuations capturing short-lived, context-dependent variations.
To this end, we propose the Low-Rank Sparse Emotion Understanding Framework (\name), a unified plug-and-play framework grounded in the Low-Rank Sparse Principle, which theoretically reframes affective dynamics as a hierarchical low-rank sparse compositional process.
Specifically, \name introduces three complementary modules: the Stability Encoding Module (\MA) extracts low-rank emotional bases by emphasizing slow-varying, context-relevant patterns while suppressing noisy high-frequency artifacts; the Dynamic Decoupling Module (\MB) isolates sparse transient signals via temporal routing and orthogonalized relational propagation, preventing transient cues from contaminating stable affective representations; and the Consistency Integration Module (\MC) reconstructs multi-scale coherence between stability and reactivity, mitigating the over-smoothing of transient cues and the dilution of stable bases under hierarchical feature aggregation.
Furthermore, we propose a Rank Aware Optimization (\MD) strategy that adaptively modulates perturbation strength according to rank- and sparsity-sensitive structure, balancing gradient smoothness and sensitivity to improve training stability and generalization.
Extensive experiments on both discrete and continuous VAC benchmarks demonstrate that \name consistently improves robustness, dynamic discrimination, and cross-dataset generalization across backbones.
Our results validate the effectiveness and generality of hierarchical low-rank sparse modeling as a principled foundation for understanding affective dynamics in videos.
\end{abstract}

\begin{IEEEkeywords}
Video-based Affective Computing, low rank sparse modeling, representation learning.
\end{IEEEkeywords}

\input{sec/1_intro}
\input{sec/2_related}

\input{sec/3_method}
\input{sec/4_exper}
\input{sec/5_con}

\section*{Acknowledgments}
This work is supported by Fundamental Research Funds for the Central Universities (Grant No. JZ2025HGTB0225), Major Scientific and Technological Project of Anhui Provincial Science and Technology Innovation Platform (Grant No. 202305a12020012), and National Natural Science Foundation of China (Grant No. 62302145).

\bibliographystyle{IEEEtran}
\bibliography{main}

\end{document}

%% file: sec/1_intro.tex
\begin{figure}[t]
  \centering
    \includegraphics[width=0.48\textwidth]{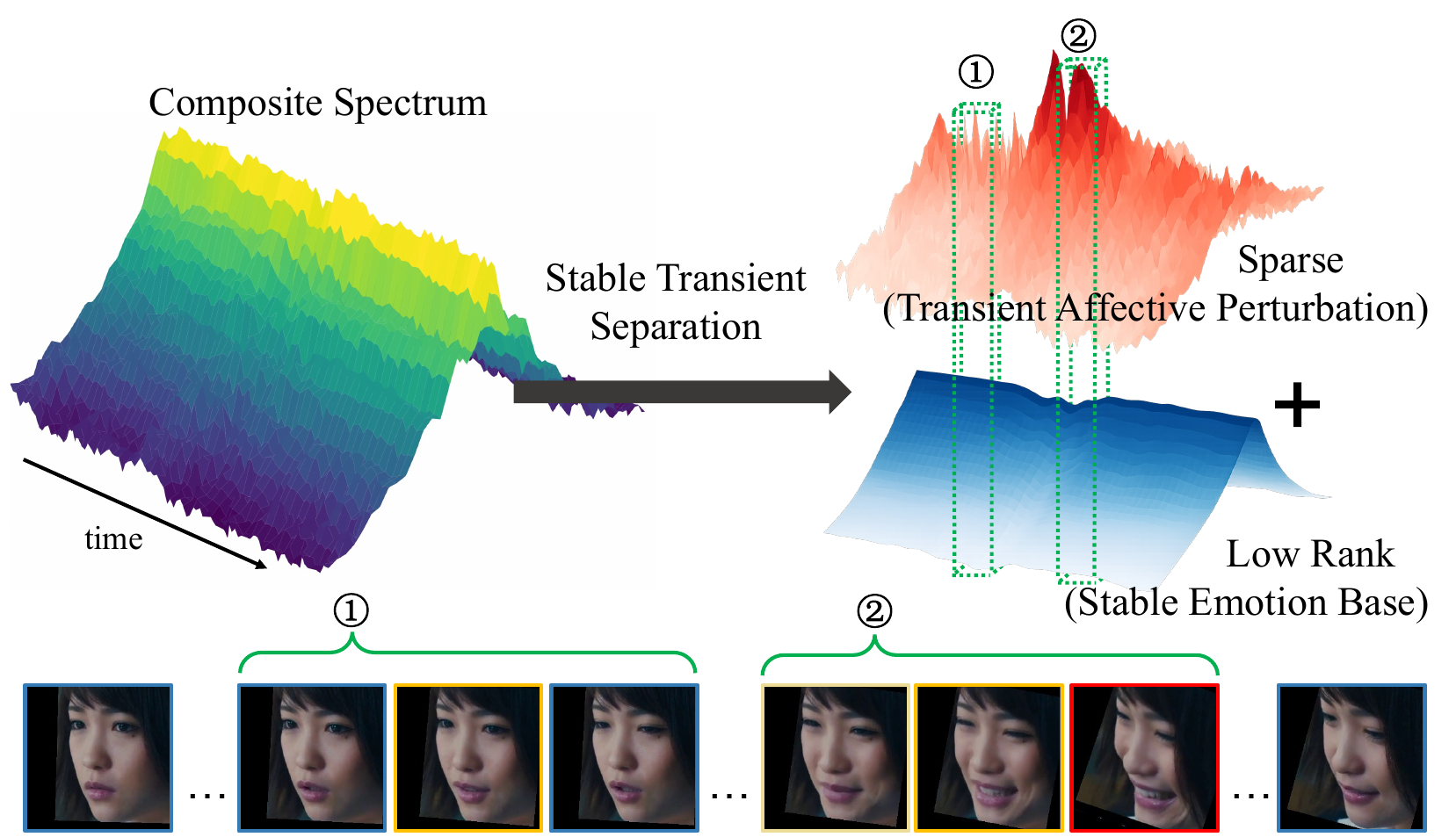}
    \caption{Illustration of the inherent low-rank sparse structure of affective dynamics. The smooth low-rank curve (\textcolor{blue}{blue}) represents stable emotional states, while the sparse curve  (\textcolor{red}{red}) highlights transient expressive surges.}
   \label{fig:onecol}
      % \vspace{-0.5cm}
\end{figure}

\section{Introduction}
\label{sec:intro}
Since showing great potential in real-world applications \cite{10879048}, \eg, psychological assessment \cite{10806792} and human–computer interaction \cite{11114734,11182317,9610134}, video-based Affective Computing (VAC) has attracted growing attention. It aims to model the dynamic evolution of human emotions from video sequences, enabling automatic perception and analysis of affective states \cite{anonymous2025every}. VAC is also increasingly relevant to consumer electronics, such as in-vehicle systems \cite{11175917} and service robots \cite{11127205}, where affect-aware interaction enhances personalization and user experience.

According to different affect representation paradigms, VAC research can be broadly categorized into two main directions, categorical emotion recognition and dimensional emotion regression \cite{11114961, 10484004}. The former aims to recognize representative emotional expressions (\eg, facial expressions \cite{10089789}) from videos to determine the overall emotion category, which is particularly useful for affective interaction and social signal recognition. The latter maps emotions onto a continuous multidimensional space, such as the valence arousal plane \cite{10848157}, to capture temporal variations in emotional intensity and psychological fluctuation, providing a more realistic characterization of continuous affect generation mechanisms.
Despite different outputs, both settings require models to preserve long-term affective trends while being sensitive to short-lived discriminative cues.
Benefiting from the rapid development of deep spatiotemporal modeling techniques, the performance of video-based affect analysis has seen significant improvements. For instance, representative studies have constructed unified 3D convolutional networks \cite{m3dfel} or spatiotemporal attention networks \cite{formerDFER, sttMa_Sun_Li_2022} to jointly capture spatial structures and temporal dynamics of video signals, thereby improving model robustness and generalization in complex real-world scenarios \cite{wang2024facialpulse}. 
However, these methods typically learn heterogeneous dynamics in a shared representation pathway, which can be fragile under non-stationary affective evidence and noisy in-the-wild videos common in consumer devices \cite{10.1145/3746027.3755036}.

Although VAC has achieved remarkable progress in both categorical recognition and dimensional regression, pioneer approaches still struggle to achieve robust and interpretable affect understanding due to the lack of structured modeling of emotional dynamics. This is because there are not only long-term emotional tones but also short-term emotional fluctuations that affect recognition results. Psychostatistical theories decompose emotions into stable components that reflect persistent psychological states and transient components driven by momentary stimuli \cite{xinli}. From the perspective of structured signal modeling, this emotional process can be characterized using a low-rank sparse framework, where long-term trends and short-term fluctuations jointly contribute to the affective dynamics \cite{dizhi}. As illustrated in Fig.~\ref{fig:onecol}, the low-rank emotional field describes smooth and persistent affective baselines, while sparse high-frequency perturbations correspond to brief, localized, and discriminative emotional variations.
This view also matches practical usage: users exhibit relatively stable baselines during a session, while key affective events appear as bursts. Without explicit disentanglement, models may overreact to spurious high-frequency artifacts or oversmooth subtle fluctuations, harming reliability in real-time applications.

However, mainstream methods typically learn these heterogeneous components jointly within a unified feature space, leading to multi-level representational entanglement and optimization instability. 
Specifically, this fundamental mechanism deficiency manifests in three key issues for visual affect analysis:
\emph{1) Structural Component Mixing:} Stable emotional trends and localized transient variations interfere with each other within the representation space, making it difficult to preserve coherent affective trajectories while capturing fine-grained changes.
\emph{2) Dynamic Propagation Coupling:} Without isolating stable and transient components during propagation, the two dynamics become cross-contaminated across spatiotemporal dimensions, weakening independent pathways for long-term tones and transient bursts.
\emph{3) Cross-scale Structural Inconsistency:} As features propagate to deeper layers and are aggregated across scales, stable trends may be diluted while sparse variations are over-smoothed, especially under efficiency-driven operations (downsampling/pooling) often used for on-device deployment.
These biases hinder maintaining global stability and local sensitivity simultaneously, resulting in unstable optimization and degraded representations.

To address these challenges, we revisit affective video modeling from the perspective of low-rank sparse emotional organization and develop the \textbf{\emph{Low-Rank Sparse Emotion understanding Framework} (\name)}, a unified architecture grounded in the low-rank sparse principle that can be seamlessly integrated into general video backbones.
\name~follows a structured dynamic hierarchical decomposition paradigm, constructing a systematic pathway from stable encoding to dynamic decoupling and consistency integration.
First, the \textbf{\emph{Stability Encoding Module} (\MA)} extracts stable emotional bases via frequency-aware decomposition with low-rank priors, suppressing random perturbations to form a coherent affective field.
Second, the \textbf{\emph{Dynamic Decoupling Module} (\MB)} decouples dynamics through temporal gating and graph orthogonalization, disentangling sparse transient signals to improve discriminability and generalization.
Third, the \textbf{\emph{Consistency Integration Module} (\MC)} aligns stability and reactivity via multi-scale fusion and relational aggregation, mitigating scale drift and over-smoothing.
Finally, \textbf{\emph{Rank Aware Optimization} (\MD)} adaptively regulates perturbations to balance low-rank smoothness and sparse sensitivity, improving training stability and generalization under heterogeneous and noisy affective patterns relevant to consumer videos.

Our main contributions are summarized as follows:
\begin{itemize}
\item To the best of our knowledge, this paper is the first to introduce the low-rank sparse principle into VAC and propose the \name framework, reformulating affective dynamics as a low-rank sparse compositional process and providing a unified lens bridging stability and sensitivity for both categorical and dimensional affect tasks.
\item We design hierarchical plug-and-play modules \MA, \MB, and \MC to enable structured decomposition, independent propagation, and multi-scale integration, addressing component entanglement, propagation coupling, and scale imbalance.
\item We propose a \MD~strategy that regulates perturbations in a rank- and sparsity-aware manner, offering an effective trade-off between stability and adaptability during optimization.
\item Extensive experiments on multiple video affective benchmarks demonstrate that \name~achieves SOTA performance in both classification and regression, with superior robustness and generalization, indicating practical potential for affect-aware consumer electronics systems.
\end{itemize}

%% file: sec/2_related.tex
\section{Related Work}
\subsection{Video-based Affective Computing}

VAC aims to understand human emotions from dynamic video sequences by modeling temporal and structural variations in affective expressions.
According to the affective representation paradigm, VAC research can be categorized into Discrete and Continuous VAC tasks \cite{ zhang2023multimodal}.

Discrete VAC focuses on mapping facial and related cues to a finite set of emotional categories (\eg, happiness, anger, sadness).
Early studies relied on handcrafted features \cite{4160945}, which were later replaced by CNN--RNN frameworks integrating spatial extraction and temporal modeling \cite{ 9102419}.
The emergence of 3D CNNs \cite{c3d} and Transformer architectures \cite{formerDFER, sttMa_Sun_Li_2022} further advanced spatiotemporal representation, enabling models to capture complex dynamics more effectively.
In addition, recent works explored multi-instance learning, part-based modeling, temporal attention, and cross-clip aggregation to mitigate sparse evidence issues, where only a few frames provide discriminative cues while the rest are neutral or noisy \cite{m3dfel}.
To address practical challenges such as sample heterogeneity, label noise, and domain shift, a growing line of research incorporated robust optimization, uncertainty-aware learning, and weak supervision \cite{10.1145/3746027.3755036, 11094811, chen2025static}, improving model stability and generalization.
Nevertheless, despite stronger backbones and training recipes, most existing methods still treat affective dynamics as a unified latent process and propagate features through coupled pathways.
This design tends to mix slowly-varying affective trends with short-lived bursts, leading to representational aliasing, sensitivity to nuisance perturbations, and limited interpretability when handling rapidly changing affective patterns.

Continuous VAC, based on dimensional emotion theory, projects emotions into a continuous valence--arousal space and models temporal evolution via regression.
Mainstream approaches employ recurrent structures or temporal Transformers \cite{zhang2023multimodal} to capture long-term dependencies and smooth affect trajectories.
However, continuous VAC is often challenged by non-stationary dynamics: affective states may drift slowly over time, but can also exhibit abrupt transitions triggered by momentary stimuli.
Purely sequential modeling without structural decomposition often over-smooths transient shifts or becomes sensitive to short-term noise, resulting in trajectory jitter and optimization instability.
Moreover, continuous regression is typically more sensitive to cross-subject and cross-device variability and thus benefits from representations that explicitly preserve global context while remaining responsive to salient local changes.
Overall, a key limitation persisting through these advances is the lack of a structured modeling principle that explicitly disentangles stable emotional baselines and transient fluctuations, and further ensures their independent yet coordinated propagation across depth and scale.

\begin{figure*}[t]
  \centering
   \includegraphics[width=0.97\linewidth]{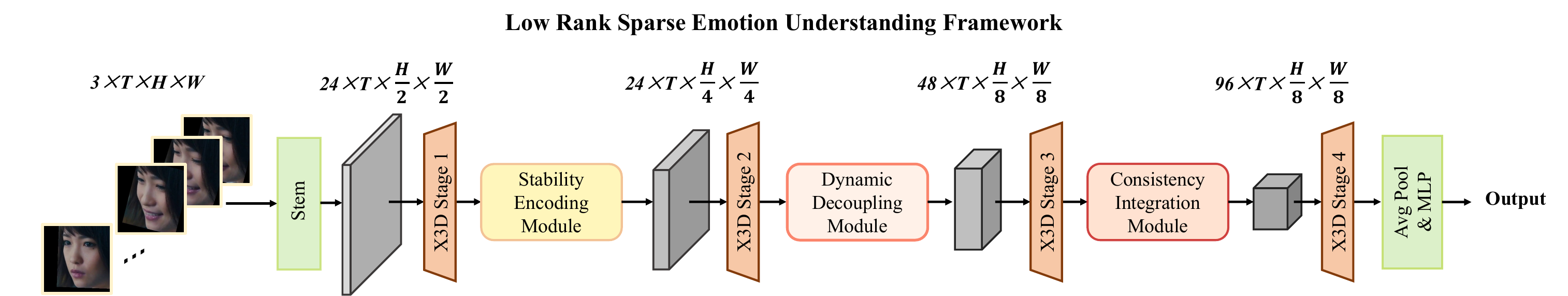}

   \caption{An overview of the proposed Low-Rank Sparse Emotion Understanding Framework (\name).}
   \label{ours}
\end{figure*}

\subsection{Low-Rank Sparse Modeling for Emotion Representation Learning}

Low-rank sparse modeling originates from high-dimensional statistics and signal decomposition theory, aiming to separate complex observations into a low-rank global structure and a sparse local perturbation for structured information extraction and noise suppression. 
Robust principal component analysis (RPCA) \cite{candes2011rpca} first formalized this concept by representing data as the sum of a low-rank matrix and a sparse matrix, offering a theoretically grounded decomposition for separating global trends from outliers.
Under incoherence conditions, this decomposition ensures identifiability and robustness to noise~\cite{bertsimas2023slr}, which makes it a natural choice for modeling signals with stable backgrounds and bursty events.
Beyond classical RPCA, subsequent works in sparse representation and structured regularization further extended low-rank and sparsity priors to dynamic settings, where temporal continuity and spatial locality can be jointly exploited for more reliable decomposition.

This principle has been widely applied to video representation learning, motion segmentation, and dynamic scene analysis, showing strong generalization and stability~\cite{zhang2023sparsedgcnn}. 
In these applications, low-rank structures often correspond to slowly changing scene content or consistent motion patterns, while sparse components capture salient events, local motion discontinuities, or rare but informative observations.
Recent studies extended this idea to deep models via low-rank attention, low-rank factorization, and sparse graph modeling, which reduce redundancy while preserving expressive power~\cite{yi2025vacsurvey}.

From a theoretical viewpoint, the low-rank sparse principle naturally aligns with the generative nature of affective signals.
Psychostatistical and affect theories suggest that emotional expressions consist of a stable affective baseline and sparse transient bursts \cite{xinli}, corresponding to the low-rank and sparse components, respectively. 
Affective signals also exhibit temporal sparsity and spatial locality, since emotions are often activated only in a few key frames or localized facial regions~\cite{lin2021sparsesampling}.
Therefore, low-rank sparse modeling provides an interpretable mechanism to preserve global affective context (stability) while amplifying discriminative short-lived cues (reactivity), rather than forcing a single latent space to explain both.
In addition, its inherent robustness can mitigate common nuisances in real-world videos, such as background clutter, illumination variations, and subject-dependent expression styles~\cite{zhang2023sparsedgcnn}.
% Despite this strong theoretical suitability, low-rank sparse modeling has not been systematically established as a unified framework for VAC that supports both discrete recognition and continuous regression, nor has it been tightly coupled with an optimization mechanism that respects rank--sparsity-sensitive structure during training.
Overall, the low-rank sparse principle offers a theoretically grounded and structurally interpretable foundation for VAC, motivating our \name~to explicitly disentangle, propagate, and integrate stable and transient affective dynamics in a unified and robust manner.

%% file: sec/3_method.tex
\section{Methodology}

We propose the \name, a unified and interpretable framework for video-based affective representation learning. 
As shown in Fig.~\ref{ours}, \name~extends X3D~\cite{Feichtenhofer_2020_CVPR} with a hierarchical pathway from stability encoding to dynamic decoupling, consistency integration, and rank aware optimization. 
\MA~ first extracts low-rank affective bases that capture invariant emotional structures and suppress noise, forming a coherent global field. Then, \MB~disentangles sparse temporal perturbations from the stable component, enabling independent propagation of transient affective cues. Next, \MC~fuses stable and dynamic representations through multiscale consistency alignment, achieving low-rank sparse equilibrium between global stability and local sensitivity. Finally, \MD~adaptively regulates gradient perturbations according to the rank sparsity balance, which ensures stable and generalizable optimization. 
Totally, these modules form a theoretically consistent and practically robust framework to jointly modeling steady affective states and different fine-grained temporal fluctuations across various heterogeneous video conditions.

\begin{figure}[t]
  \centering
   \includegraphics[width=0.85\linewidth]{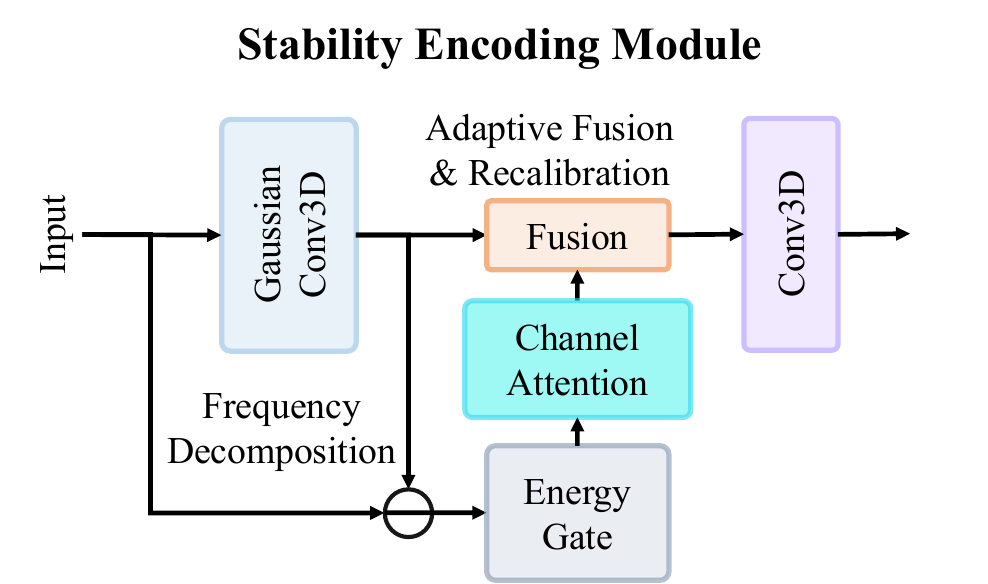}
   \caption{The Stability Encoding Module.}
   \label{ours2}
\end{figure}

\subsection{Stability Encoding Module}

As shown in Fig. \ref{ours2}, the goal of \MA~is to construct a structurally consistent emotional base by emphasizing low-rank temporal trends while regularizing transient variations, thereby providing a clean foundation for subsequent dynamic decoupling.
This module is designed to (i) stabilize affective trajectories by extracting slow-varying components that encode global emotional context, and (ii) preserve informative short-term dynamics without amplifying noise-like high-frequency artifacts.
Given the input video feature $\mathcal{X} \in \mathbb{R}^{B \times C \times T \times H \times W}$, where $B$ is batch size, $C$ is channel dimension, and $(T,H,W)$ denote temporal length and spatial resolution, we decompose $\mathcal{X}$ into two complementary components:
\begin{equation}
\mathcal{X} = \mathcal{X}_{\text{low}} + \mathcal{X}_{\text{high}},
\end{equation}
where $\mathcal{X}_{\text{low}}$ denotes the low-rank structural component capturing stable emotional trends (long-term tone), and $\mathcal{X}_{\text{high}}$ represents sparse transient perturbations corresponding to short-lived expressive changes.
This decomposition provides an explicit separation between stability-oriented and reactivity-oriented signals, making the subsequent modeling stages more controllable and interpretable.

To realize this separation in a differentiable and efficient manner, we perform Gaussian-based frequency decomposition:
\begin{equation}
\mathcal{X}_{\text{low}} = \mathcal{G}_k * \mathcal{X},\quad 
\mathcal{X}_{\text{high}} = \mathcal{X} - \mathcal{X}_{\text{low}},
\end{equation}
where $\mathcal{G}_k$ is a 3D Gaussian kernel with window size $k$ and $*$ denotes 3D convolution. 
As a principled low-pass operator, $\mathcal{G}_k$ extracts temporally smooth and spatially coherent tendencies, which align with low-rank affective bases under the assumption that stable emotional context varies slowly and exhibits strong redundancy across frames.
The residual branch $\mathcal{X}_{\text{high}}$ captures high-frequency variations, including both meaningful transient affect cues (e.g., brief muscle activations) and nuisance perturbations (e.g., illumination flicker, camera jitter, compression artifacts). 
Therefore, instead of naively suppressing $\mathcal{X}_{\text{high}}$, we introduce a selective refinement mechanism to retain informative transients while attenuating noise-like fluctuations.

Specifically, we apply energy modulation and channel attention:
\begin{equation}
\tilde{\mathcal{X}}_{\text{high}} = \sigma(\text{CA}(\mathcal{X}_{\text{high}})) \odot \mathcal{E}(\mathcal{X}_{\text{high}}),
\end{equation}
where $\mathcal{E}(\cdot)$ denotes depthwise energy gating, $\text{CA}(\cdot)$ is channel attention, $\sigma(\cdot)$ is the sigmoid function, and $\odot$ denotes element-wise multiplication.
Intuitively, $\mathcal{E}(\cdot)$ performs a lightweight dynamic response filtering: channels/locations with weak or noisy activations are suppressed, while salient responses are preserved to maintain sensitivity to transient affective bursts.
Meanwhile, $\text{CA}(\cdot)$ reweights channels according to their affect relevance, encouraging the refined high-frequency branch to focus on semantically meaningful dynamics rather than background variations.

Finally, the low- and high-frequency representations are adaptively fused:
\begin{equation}
\mathcal{X}_{\text{fused}} = \lambda \cdot \mathcal{X}_{\text{low}} + (1 - \lambda) \cdot \tilde{\mathcal{X}}_{\text{high}},
\end{equation}
where $\lambda \in (0,1)$ is a learnable fusion coefficient that controls the balance between low-rank stability and sparse sensitivity.
This adaptive fusion is crucial because the relative importance of stable context and transient cues varies across videos, emotion categories, and affect intensity: for slowly evolving affect, a larger $\lambda$ stabilizes trajectories; for highly dynamic segments, $(1-\lambda)$ allows transient cues to contribute more effectively.

Through frequency-aware decomposition and selective refinement, \MA~produces a structurally coherent emotional base that preserves essential affective dynamics while suppressing disruptive noise, providing a reliable foundation for subsequent dynamic decoupling and cross-scale consistency modeling.

\begin{figure}[t]
  \centering
   \includegraphics[width=0.85\linewidth]{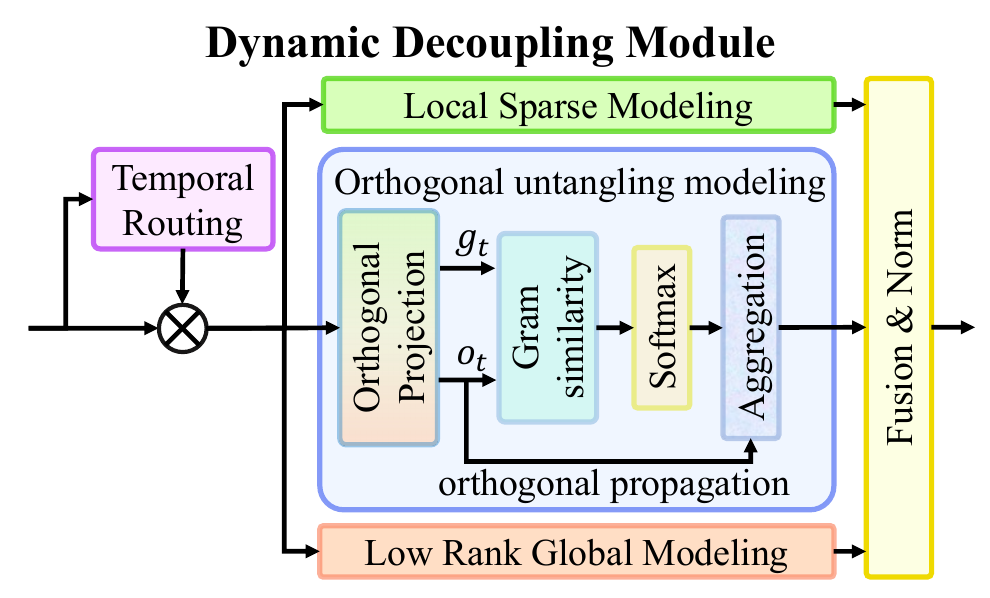}

   \caption{The Dynamic Decoupling Module.}
   \label{ours3}
\end{figure}

 \subsection{Dynamic Decoupling Module}

Building upon the stable emotional bases provided by the \MA, the \MB~further processes the feature tensor to achieve explicit separation between sparse temporal perturbations and stable components. 
As shown in Fig. \ref{ours3}, the key idea is to construct a decoupled propagation pathway: stable bases should remain globally coherent, while transient cues should be selectively routed and propagated without contaminating the stable manifold.
Concretely, \MB~consists of a temporal routing gate that filters and reweights temporal evidence under an information bottleneck principle, and a spatial relational interaction block that enforces approximate orthogonality between propagation subspaces to reduce cross-contamination.

We first construct a temporal routing gating mechanism that achieves adaptive temporal weighting through information bottleneck principles:
\begin{equation}
\mathcal{G}_t = \sigma \left( \mathbf{W}_2 \cdot \delta \left( \mathbf{W}_1 \cdot \mathcal{P}(\mathcal{X}) \right) \right),
\end{equation}
where $\mathcal{P}(\cdot)$ denotes spatial average pooling, which compresses spatial redundancy and yields a compact temporal descriptor that reflects how affective evidence evolves over time.
$\mathbf{W}_1, \mathbf{W}_2 \in \mathbb{R}^{T \times T}$ are fully-connected projection matrices, $\delta$ is ReLU, and $\sigma$ is sigmoid. 
The gate $\mathcal{G}_t \in (0,1)^{B \times 1 \times T \times 1 \times 1}$ (after proper reshaping/broadcasting) acts as a soft temporal selector: it suppresses uninformative or noisy frames while emphasizing key moments that carry discriminative transient affective evidence.
The temporally routed representation is obtained via Hadamard product: $\tilde{\mathcal{X}} = \mathcal{X} \odot \mathcal{G}_t$.
Importantly, this routing is \emph{content-adaptive}: rather than assuming uniform relevance across time, it learns to allocate representation capacity to short-lived affective bursts, which are common in in-the-wild videos and are critical for dynamic affect understanding.

To further decorrelate local semantic patterns from dynamic responses, \MB~constructs a spatial graph interaction system based on manifold orthogonalization:
\begin{equation}
\hat{\mathcal{G}} = \mathcal{L}_2(\Phi_g(\tilde{\mathcal{X}})), \quad
\hat{\mathcal{O}} = \mathcal{L}_2(\Phi_o(\tilde{\mathcal{X}})), 
\end{equation}
\begin{equation}
\mathcal{X}_{\text{graph}} = \mathcal{O} \otimes \text{Softmax}\left( \hat{\mathcal{G}}^\top \hat{\mathcal{O}} \right),
\end{equation}
where $\Phi_g, \Phi_o$ are independent convolutional projection operators and $\mathcal{L}_2$ denotes L2 normalization.
Let $N=H\times W$ be the number of spatial positions; the similarity $\hat{\mathcal{G}}^\top \hat{\mathcal{O}} \in \mathbb{R}^{N \times N}$ builds a relational map between spatial nodes, and the Softmax normalizes it into propagation weights.
Due to L2 normalization, the inner-product similarity effectively measures angular consistency, which encourages stable and transient channels to occupy different directions in the representation manifold.
This design enforces approximate orthogonality between graph channels, thereby reducing redundancy and preventing transient activations from being diffusely propagated into the stable pathway (and vice versa).
From a geometric standpoint, this yields a more disentangled propagation behavior on the Riemannian manifold induced by normalized features, improving interpretability and robustness.

Finally, \MB~integrates three complementary feature subspaces through tensor concatenation and convolution:
\begin{equation}
\mathcal{X}_{\text{out}} = \Psi_{1\times1}\left( \left[ \mathcal{X}_{\text{local}}, \mathcal{X}_{\text{graph}}, \mathcal{X}_{\text{global}} \right] \right),
\end{equation}
where $\mathcal{X}_{\text{local}} = \Theta_{\text{local}}(\tilde{\mathcal{X}})$ represents local dynamic features preserving fine-grained, spatially localized affect cues, and $\mathcal{X}_{\text{global}} = \Theta_{\text{global}}(\tilde{\mathcal{X}})$ denotes global temporal context features capturing long-range affective dependencies.
$\Psi_{1\times1}$ adaptively fuses them and restores channel capacity. 
This multi-subspace design provides complementary signals: local captures micro-variations, graph captures relational propagation under orthogonalized constraints, and global maintains context consistency.
Through temporal routing, manifold orthogonalization, and multi-subspace fusion, \MB~progressively decouples stable trends from transient dynamic variations, ensuring independent and interpretable propagation pathways while producing structurally coherent dynamic features for subsequent multi-scale consistency integration.

\begin{figure}[t]
  \centering
   \includegraphics[width=0.90\linewidth]{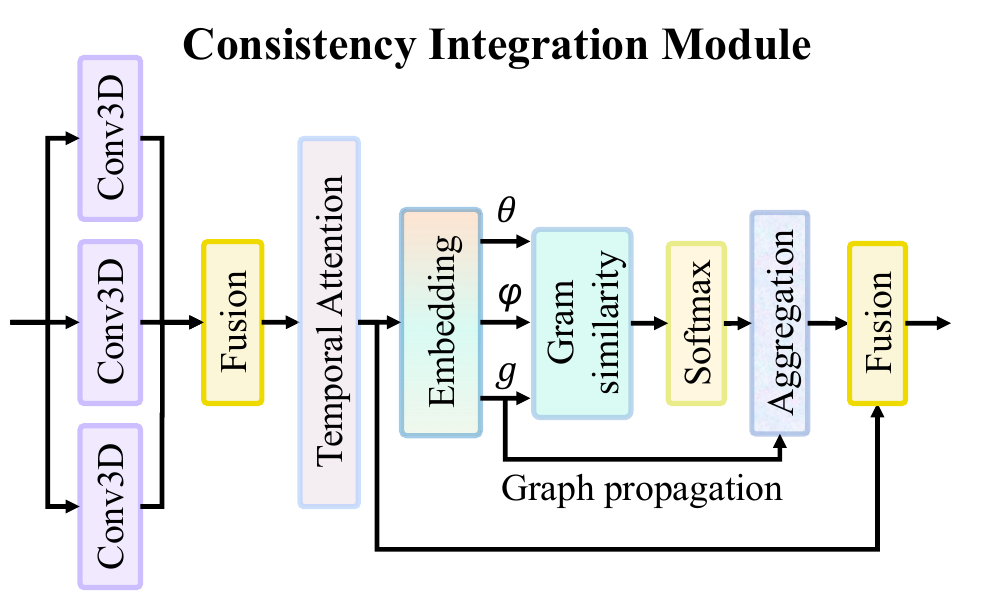}

   \caption{The Consistency Integration Module.}
   \label{ours4}
\end{figure}

\subsection{Consistency Integration Module}

As shown in Fig. \ref{ours4}, \MC~aims to achieve cross-scale consistency alignment between the low-rank stable field and sparse dynamic representations based on the decoupling from preceding modules, constructing hierarchical and structurally balanced spatiotemporal affective representations.
Therefore, \MC~explicitly coordinates multi-scale structural cues, temporal consistency, and relational dependency aggregation to maintain a low-rank sparse organization throughout depth.

First, we construct a family of multi-scale depthwise separable operators $\{\Psi_{s_i}\}_{i=1}^K$ that capture structural dependencies from fine to coarse granularity through variable scale spatial kernels:
\begin{equation}
\mathcal{X}_{\text{ms}} = \Phi_{\text{fuse}}\left( \bigoplus_{i=1}^{K} \Psi_{s_i}(\mathcal{X}) \right),
\end{equation}
where $\oplus$ denotes channel concatenation and $\Phi_{\text{fuse}}$ is a fusion convolution.
Depthwise separable operators provide an efficient way to enrich receptive fields while controlling computation, which is also favorable for deployment-oriented settings.
The multi-scale family $\{\Psi_{s_i}\}$ encourages the representation to preserve both local discriminative patterns (small kernels) and global affective layouts (large kernels), thereby improving the structural completeness of affective cues.

To enhance cross-frame emotional consistency, \MC~introduces a temporal attention recalibration mechanism based on global affective consistency:
\begin{equation}
\mathcal{X}_{\text{ta}} = \mathcal{X}_{\text{ms}} \odot \sigma\left( \Gamma_{\text{temp}}(\mathcal{X}_{\text{ms}}) \right),
\end{equation}
where $\Gamma_{\text{temp}}(\cdot)$ is a composite operator of temporal pooling and sequential attention mapping.
This mechanism acts as a consistency-aware temporal filter: it strengthens temporally stable affect segments that should dominate the emotional bases, while still allowing sharp transient variations to pass when they are consistently supported by the multi-scale representation.
In practice, this helps reduce jitter in predicted trajectories and prevents unstable bursts caused by nuisance perturbations.

Finally, we construct a non-local graph relational modeling operator that achieves cross-scale dependency aggregation through feature projection and relational weighting:
\begin{equation}
\mathcal{A} = \text{Softmax}\left( \frac{\Theta(\mathcal{X}_{\text{ta}})^\top \Phi(\mathcal{X}_{\text{ta}})}{\sqrt{C}} \right),
\end{equation}
\begin{equation}
\mathcal{X}_{\text{out}} = \Gamma_{\text{graph}}(\mathcal{A}, \mathcal{X}_{\text{ta}}),
\end{equation}
where $\Theta$ and $\Phi$ are projection operators and $\Gamma_{\text{graph}}$ is the graph aggregation operator.
The affinity matrix $\mathcal{A}$ establishes long-range relational consistency among spatiotemporal nodes, allowing distant but semantically consistent affect evidence to reinforce each other.
This is particularly beneficial when transient cues are spatially localized or temporally short-lived: relational aggregation helps recover coherent semantics by linking them to supportive contexts across space and time.
By jointly integrating multi-scale structural cues, global temporal consistency, and relational graph modeling, \MC~achieves a representational equilibrium between low-rank stability and sparse dynamic sensitivity, yielding a unified affective embedding that is globally coherent yet locally responsive.

\subsection{Rank Aware Optimization}

To achieve stable yet adaptive optimization under heterogeneous affective representations, we propose the \MD, constructing a gradient perturbation mechanism based on structural awareness.
Unlike traditional homogeneous perturbation methods~\cite{foret2021sharpnessaware}, \MD~performs hierarchical perturbation modulation by explicitly analyzing intrinsic low-rank and sparse characteristics of parameter tensors.
The motivation is that different parameter groups contribute differently to stability and sensitivity; thus, using a uniform perturbation radius can be suboptimal and may leave residual directional instability.

For the parameter tensor $\mathcal{W} \in \mathbb{R}^{d_1 \times d_2 \times \cdots \times d_k}$, we define structural sensitivity metrics:
\begin{equation}
\rho_r = \frac{\|\Sigma(\mathcal{W}) > \tau_r\|_0}{\text{dim}(\Sigma(\mathcal{W}))}, \quad 
\rho_s = \frac{\| |\mathcal{W}| < \tau_s \|_0}{|\mathcal{W}|},
\end{equation}
where $\rho_r$ measures rank sensitivity by the proportion of singular values exceeding threshold $\tau_r$, and $\rho_s$ measures sparsity sensitivity by the proportion of near-zero entries under threshold $\tau_s$.
Intuitively, a larger $\rho_r$ implies a parameter tensor with richer effective rank, where stronger perturbation can encourage flatter minima and better generalization; a larger $\rho_s$ indicates high sparsity, where overly aggressive perturbation may destabilize sensitive sparse structures and harm convergence.
Based on these metrics, we define the dynamic perturbation radius:
\begin{equation}
\rho_{\text{dyn}} = \rho_{\text{base}} \cdot \left( 1 + \alpha \cdot \rho_r - \beta \cdot \rho_s \right),
\end{equation}
where $\rho_{\text{base}}$ is the base perturbation coefficient and $\alpha,\beta$ control the trade-off between low-rank expansion and sparsity suppression.
This formulation yields an explicit special case interpretation: when $\alpha=\beta=0$, \MD~reduces to a homogeneous perturbation scheme with constant radius $\rho_{\text{base}}$; when $\alpha>0$ and $\beta>0$, the perturbation becomes structure-adaptive, strengthening perturbations on rank-rich components while constraining perturbations on sparse-sensitive regions.

\begin{algorithm}[t]
\caption{Rank Aware Optimization}
\label{algo:rao}
\KwIn{Parameters $\mathcal{W}$, learning rate $\eta$, base optimizer $\mathcal{O}$, base perturbation $\rho_{\text{base}}$, coefficients $\alpha,\beta,\gamma$}
\KwOut{Optimized parameters $\mathcal{W}$}
\BlankLine
\ForEach{parameter tensor $\mathcal{W}$}{
    Estimate structural sensitivities: $\rho_r, \rho_s \leftarrow \Gamma_{\text{struct}}(\mathcal{W})$\;
    Compute dynamic radius: $\rho_{\text{dyn}} \leftarrow \rho_{\text{base}} \cdot (1 + \alpha\rho_r - \beta\rho_s)$\;

    Normalize gradient: $\mathcal{G} \leftarrow \nabla_{\mathcal{W}} \mathcal{L} / \|\nabla_{\mathcal{W}} \mathcal{L}\|_F$\;
    Apply perturbation: $\mathcal{W}' \leftarrow \mathcal{W} + \rho_{\text{dyn}} \cdot \mathcal{G}$\;
    Compute perturbed loss: $\mathcal{L}' = \mathcal{L}(\mathcal{W}')$\;

    Backward update: $\mathcal{W} \leftarrow \mathcal{W} - \eta \nabla_{\mathcal{W}'} \mathcal{L}'$\;
    Adaptive adjustment: $\rho_{\text{base}} \leftarrow \rho_{\text{base}} \cdot (1 + \gamma \cdot \delta)$\;
}
\Return{$\mathcal{W}$}\;
\end{algorithm}

Subsequently, \MD~achieves structure-driven learning through a dual-phase perturb-and-update process, as summarized in Alg.~\ref{algo:rao}.
The optimizer synergizes structure-aware perturbation with feedback adaptation: rank-dominant components receive stronger perturbations to encourage flatter solutions and improve generalization, while sparse-sensitive regions are constrained to prevent brittle updates and maintain convergence stability.
This balancing mechanism is particularly suitable for VAC, where heterogeneous affective dynamics can induce non-uniform curvature and optimization sensitivity across parameter groups.

Through \MD, \name\ achieves stable and structure-aware learning dynamics, ensuring that the low-rank sparse organization remains consistent throughout optimization and enabling reliable modeling of complex affective temporal patterns across both categorical and dimensional VAC tasks.

%% file: sec/4_exper.tex
\begin{table*}[t]
\renewcommand{\arraystretch}{1.2}
\centering
\setlength{\tabcolsep}{3mm}
\small

\scalebox{1.0}{
\begin{tabular}{ c|c|ccccccc|cc }
\hline
\multirow{2}{*}{Method} & \multirow{2}{*}{years} & \multicolumn{7}{c|}{Accuracy of Each Emotion(\%)} & \multicolumn{2}{c}{Metrics (\%)}  \\ 
\cmidrule(lr){3-11}
& & Hap.	  & Sad.    & Neu.   & Ang.	   & Sur.	& Dis.	  & Fea.	 & WAR   	& UAR \\ \hline
 
ResNet18+LSTM \cite{dfew} &ACM MM'20 &78.00 &40.65 &53.77 &56.83 &45.00 &4.14 &21.62 &53.08 &42.86 \\
EC-STFL \cite{dfew}  &ACM MM'20 &79.18 &49.05 &57.85 &60.98 &46.15 &2.76 &21.51 &56.51 &45.35  \\
Former-DFER \cite{formerDFER} &ACM MM'21  &84.05	 &62.57	   &67.52	&70.03	  &56.43	&3.45	 &31.78	    &65.70	   &53.69 \\ 
STT \cite{sttMa_Sun_Li_2022}   &arXiv'22     &87.36	 &67.90   &64.97	&71.27	  &53.10	&3.49	 &34.04	    &66.45	   &54.58 \\ 
Logo-Former\cite{Ma2023LogoFormerLS} & ICASSP'23 &85.39 &66.52 &68.94 &71.33 &54.59 &0.00 &32.71 &66.98  &54.21   \\
NR-DFERNet \cite{nrdfernetnoiserobustnetworkdynamic} &arXiv'22   &88.47	 &64.84	   &70.03	& 75.09 &61.60	&0.00	 &19.43	    &68.19	   &54.21 \\

GCA+IAL \cite{ial/aaai.v37i1.25077}   &AAAI'23         &87.95	 &67.21	   &\underline{70.10}	&76.06	  &62.22	&0.00	 &26.44	    &69.24	   &55.71\\ 
M3DFEL \cite{m3dfel} &CVPR'23   &89.59 	 &68.38   &67.88	&74.24	  &59.69	&0.00	 &31.63  	&69.25	   &56.10 \\  
T-MEP \cite{10250883} &T-CSVT'24 &N/A&N/A&N/A&N/A&N/A&N/A&N/A&  68.85& 57.16
\\
LG-DSTF \cite{Zhang_Tian_Zhang_Guo_Xu} &T-MM'24&N/A&N/A&N/A&N/A&N/A&N/A&N/A& \underline{69.82} & 58.89\\
CFAN-SDA \cite{Chen_Wen_Yang_Li_Chen_Wang} &T-CSVT'24  &\textbf{90.84} &\textbf{70.91} &65.72 &69.97 &57.86 &13.10 &\underline{35.36} &69.19& 57.70\\
% CLIPER \cite{10687508}  &ICME'24 &N/A&N/A&N/A&N/A&N/A&N/A&N/A&  \underline{70.84}& 57.56 \\
RDFER\cite{10908623} & T-BIOM'25 &89.69 	 &69.22  &\textbf{70.18}	&71.47	&62.08	&0.69	 &28.71  	&69.73	   &56.93  \\ 
ST-RDGCN \cite{10844531} &T-AFFC'25  &89.57 &69.92  &62.17 &\textbf{79.31} & \underline{62.24} & \textbf{20.69} &30.39 &   69.37 & \underline{59.18} \\
% HDF\cite{10.1145/3746027.3755036}   &ACM MM'25 &89.67 &\textbf{71.20}	&67.42	&73.03	&\underline{64.44}	&12.41	&\textbf{41.63}	&\underline{71.60}	&\underline{60.40} \\ 
\cmidrule(lr){1-11}
% \name~(Ours)   &- &87.53 &66.23 &69.85 &79.26 &63.27 &10.34 &46.67 &71.36&60.45 \\
% \name~(Ours)   &- &89.78 &71.77 &66.48 &70.57 &67.69 &20.69 &28.18 &70.87&59.31 \\
% \name~(Ours)   &- &89.16 &70.90 &72.98 &74.19 &71.67 &3.45  &32.78 &72.13 &60.88 \\
% \name~(Ours)   &- &91.20 &71.16 &64.73 &72.35 &68.26 &31.03 &37.22 &71.96&62.28 \\
% \name~(Ours)   &- &88.18 &69.58 &56.09 &82.95 &66.55 &24.14 &51.11 &72.22 &62.65\\
\name~(Ours)   &- &\underline{89.77} &\underline{69.93} &66.03 &\underline{76.86} &\textbf{67.49} &\underline{17.93} &\textbf{39.19}&\textbf{71.71}&\textbf{61.12} \\\hline
\end{tabular}
}

\caption{Comparison (\%) of our \name~with the SOTA methods on DFEW 5-fd (\textbf{Bold}: Best, \ul{Underline}: Second best).}

\label{res-dfew}
%  \vspace{-0.4cm}
\end{table*}

\section{Experiments}
\subsection{Experimental Setup}
\subsubsection{Datasets}
 
We conduct experiments on three representative in-the-wild video affective computing benchmarks, including two categorical datasets DFEW \cite{dfew} and FERV39k \cite{ferv39k} and one dimensional dataset VEATIC \cite{10484004}. 

DFEW and FERV39k are large-scale categorical datasets collected from real-world movies and diverse scenes, covering seven basic emotion categories, happy, sad, neutral, angry, surprised, disgusted, and fearful. DFEW contains over 16 000 clips from more than 1 500 films with significant variations in illumination and pose, while FERV39k extends to 38 935 clips across four major scenes and 22 subdomains, annotated by 30 professional raters to ensure label reliability.
For continuous emotion regression, we adopt VEATIC, a video-based valence–arousal dataset that captures fine-grained temporal variations of affect under unconstrained conditions, enabling evaluation of our framework's generalization from categorical recognition to continuous affect prediction.

\subsubsection{Metrics} 
To maintain consistency with prior works, we evaluate categorical emotion recognition using weighted average recall (WAR) and unweighted average recall (UAR), and assess continuous emotion regression with root mean square error (RMSE). WAR measures overall recognition accuracy weighted by class frequency, reflecting real world effectiveness under imbalanced emotion distributions, while UAR equally averages recall across all categories to assess balanced performance. For continuous valence arousal estimation, RMSE quantifies the deviation between predicted and ground-truth trajectories, where lower values indicate more precise modeling of fine-grained emotional dynamics.

\subsubsection{Implementation Details} 
Our entire framework is implemented using PyTorch-GPU and trained on 4 NVIDIA RTX A6000 GPUs. In our experiment, all images are resized to 112×112. The model is trained using AdamW as the base optimizer in combination with a cosine scheduler. The learning rate is set to 1e-4, with a minimum learning rate of 5e-6. 

\begin{table}[h]
\renewcommand{\arraystretch}{1.2}
\centering
\small
\setlength{\tabcolsep}{3mm}
\scalebox{1.0}{
\begin{tabular}{ c|cc }
\hline

\multirow{2}{*}{Method} & \multicolumn{2}{c}{Metrics (\%)}   \\ \cmidrule(lr){2-3}
 & WAR   & UAR \\ \hline
2C3D\cite{ferv39k} &41.77 &30.72\\
ResNet18+LSTM \cite{ferv39k}  &42.59 &30.92 \\
2ResNet18+LSTM \cite{ferv39k} &43.20 &31.28\\

VGG13+LSTM \cite{ferv39k}  &43.37 &32.42 \\
2VGG13+LSTM\cite{ferv39k}  &44.54 &32.79 \\

Former-DFER \cite{formerDFER}  &46.85 &37.20\\
M3DFEL \cite{m3dfel} &47.67	&35.94 \\

Logo-Former \cite{Ma2023LogoFormerLS} &48.13	&38.22 \\
LG-DSTF \cite{Zhang_Tian_Zhang_Guo_Xu} &48.19 &39.84  \\ 

GCA+IAL \cite{ial/aaai.v37i1.25077}  &48.54	&35.82\\
RDFER \cite{10908623} &48.60 &36.47  \\ 
CFAN-SDA \cite{Chen_Wen_Yang_Li_Chen_Wang} &49.48 & 39.56\\
ST-RDGCN \cite{10844531} &49.03 &40.48\\
HDF \cite{10.1145/3746027.3755036} &\underline{50.30}		&\underline{40.49}	\\ 
\cmidrule(lr){1-3} 
\name~(Ours)	& \textbf{50.73}	& \textbf{41.26}\\ \hline
\end{tabular}
}
\caption{Comparison (\%) of our \name~with the SOTA methods on FERV39k.}
\label{res-39k}
%  \vspace{-0.4cm}
\end{table}

\begin{table}[h]
\renewcommand{\arraystretch}{1.2}
\centering
\setlength{\tabcolsep}{3mm}
\scalebox{1.0}{
\begin{tabular}{ c|cc cc }
\hline
\multirow{2}{*}{Method} & \multicolumn{3}{c}{RMSE}   \\ \cmidrule(lr){2-4}
  & Valence & Arousal& Overall\\ \hline
EMOTIC \cite{Kosti_Alvarez_Recasens_Lapedriza_2019}  &0.3219 &0.2645 &0.2931  \\ 
X3D \cite{10.1145/3746027.3755036} & 0.3136 & 0.2516 & 0.2826   \\ 
VEATIC \cite{10484004}    &0.3107 &0.2453&0.2780 \\ 
\cmidrule(lr){1-4} 
\name~(Ours)	&\textbf{0.3094}		&\textbf{0.2369}	 &\textbf{0.2732}	\\ \hline
\end{tabular}}
\caption{Comparison (\%) of our \name~with the commonly used Continuous VAC methods on VEATIC.}
% \label{veatic}
%    \vspace{-0.4cm}
\end{table}

\begin{figure*}[t]
\centering
% \hspace*{-1cm}
\includegraphics[width=0.85\textwidth]{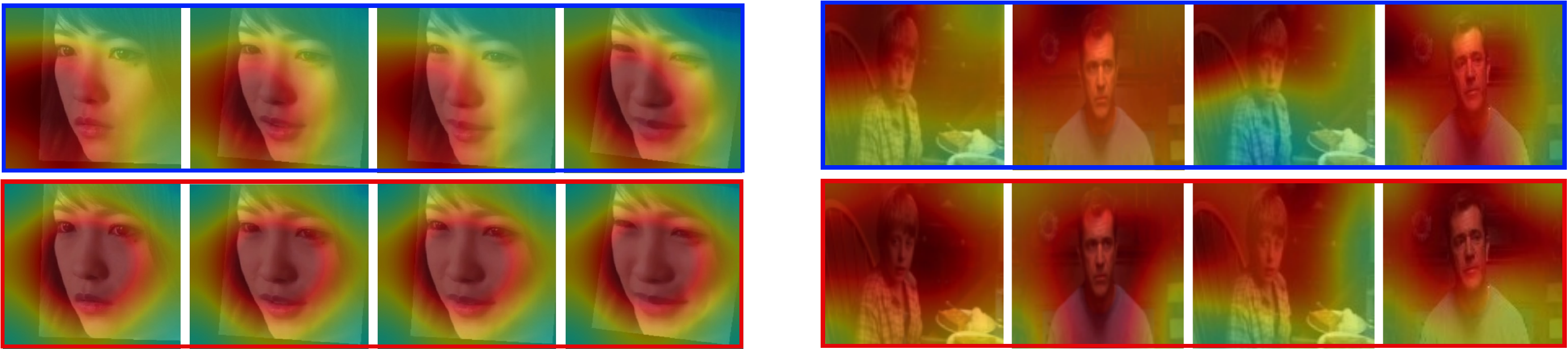} % Reduce the figure size so that it is slightly narrower than the column.
\caption{ Visualization of the learned feature maps.}

\label{vis}
% \vspace{-0.4cm}
\end{figure*}

\subsection{Comparison with the State-of-the-art Methods}

\subsubsection{Results on Discrete VAC Tasks}
We evaluate \name~on two large-scale discrete video affective categorization benchmarks, DFEW and FERV39k, under the official split protocols. As shown in Tab.~\ref{res-dfew} and Tab.~\ref{res-39k}, \name~achieves consistent SOTA performance on both datasets across WAR and UAR metrics. Compared with recent transformer-based and hybrid spatiotemporal models, \name~delivers not only stronger overall recognition accuracy (WAR) but also more balanced class-wise performance (UAR), indicating that the proposed hierarchical low-rank sparse modeling alleviates the over-dominance of majority classes and improves robustness under class imbalance.
In particular, the gains on UAR suggest that \name~better preserves discriminative cues for minority or visually subtle emotion categories, which are often overwhelmed by dominant coarse affect patterns when representations are learned in a fully entangled manner.
This effect is consistent with our motivation: stable emotional bases provide a coherent global context, while sparse transient fluctuations capture short-lived, discriminative bursts; disentangling them prevents transient cues from being washed out by global averaging and prevents stable trends from being corrupted by noisy perturbations.

% Moreover, \name~exhibits strong stability across datasets with distinct annotation styles and expression distributions. DFEW and FERV39k differ in emotion coverage, video quality, and scene diversity; nevertheless, \name~maintains consistent improvements under the same training recipe. This is particularly meaningful for practical deployment in consumer electronics settings, where capture conditions and user behaviors vary significantly across devices and environments. Remarkably, we use the same training configuration for both datasets without task-specific tuning, yet \name~achieves robust results, confirming its adaptability and scalability across heterogeneous affective domains.
Overall, these results verify that explicitly modeling affective dynamics as a hierarchical low-rank sparse composition yields more transferable and reliable video emotion understanding than purely monolithic feature learning.

\subsubsection{Results on Continuous VAC Tasks}
We further evaluate \name~on the continuous video affective regression benchmark VEATIC, where frame-level valence--arousal trajectories are estimated. As presented in Tab.~\ref{veatic}, \name~achieves the lowest RMSE on both valence and arousal dimensions, surpassing prior temporal--frequency modeling and distribution-aware regression frameworks. These improvements are particularly notable for continuous VAC because regression is highly sensitive to temporal instability: small errors in modeling transient shifts can accumulate into trajectory drift, while over-smoothing can suppress meaningful affective transitions.
In contrast, \name~explicitly separates stable emotional trends (low-rank bases) from short-term affective shifts (sparse fluctuations) and then reconstructs cross-scale coherence through consistency integration, which jointly reduces trajectory noise and preserves salient local changes.

From an affect-dynamics perspective, valence often reflects longer-term mood-related tendencies, while arousal is more reactive to momentary stimuli; the consistent RMSE reduction in both dimensions indicates that \name~can simultaneously maintain global stability and local sensitivity.
This is aligned with our framework design: \MA stabilizes affective baselines by suppressing noisy high-frequency artifacts, \MB prevents transient bursts from contaminating stable representations while preserving discriminative temporal perturbations, and \MC coordinates multi-scale aggregation to avoid diluting bases or over-smoothing fluctuations.
Overall, \name~provides a unified modeling paradigm capable of delivering robust and interpretable affective understanding across both discrete and continuous VAC scenarios, which is desirable for real-world consumer-facing affect applications requiring stable yet responsive predictions.

\begin{table}[t]
\setlength{\tabcolsep}{1.2mm}
\renewcommand{\arraystretch}{1.2}

\centering
\small
\scalebox{1.0}{
\begin{tabular}{l|cc|c}
\hline
   \makecell[c]{\multirow{2}{*}{ Method}} & \multicolumn{2}{c|}{DFEW} &\multicolumn{1}{c}{VEATIC} \\ \cmidrule(lr){2-4} 
   & WAR & UAR  &RMSE        \\ \hline
  base. & 68.12 & 58.21 & 0.2826\\\cmidrule(lr){1-4} 
   + \MA & 68.88  &  59.15  & 0.2812\\  
   + \MA~+ \MB&  70.72 & 60.57  & 0.2787 \\ 
   + \MA~+ \MB~+ \MC    &71.88 & 60.86 & 0.2753\\\cmidrule(lr){1-4} 
   + \MA~+ \MB~+ \MC~+ \MD  &\textbf{72.22} & \textbf{62.65} & \textbf{0.2732}\\\hline
\end{tabular}}
\caption{Ablation study of different components in \name~on DFEW (fd5) and VEATIC. }

\label{ablation}
   % \vspace{-0.4cm}
\end{table}

\begin{figure}[t]
\centering
% \hspace*{-1cm}
\includegraphics[width=0.3\textwidth]{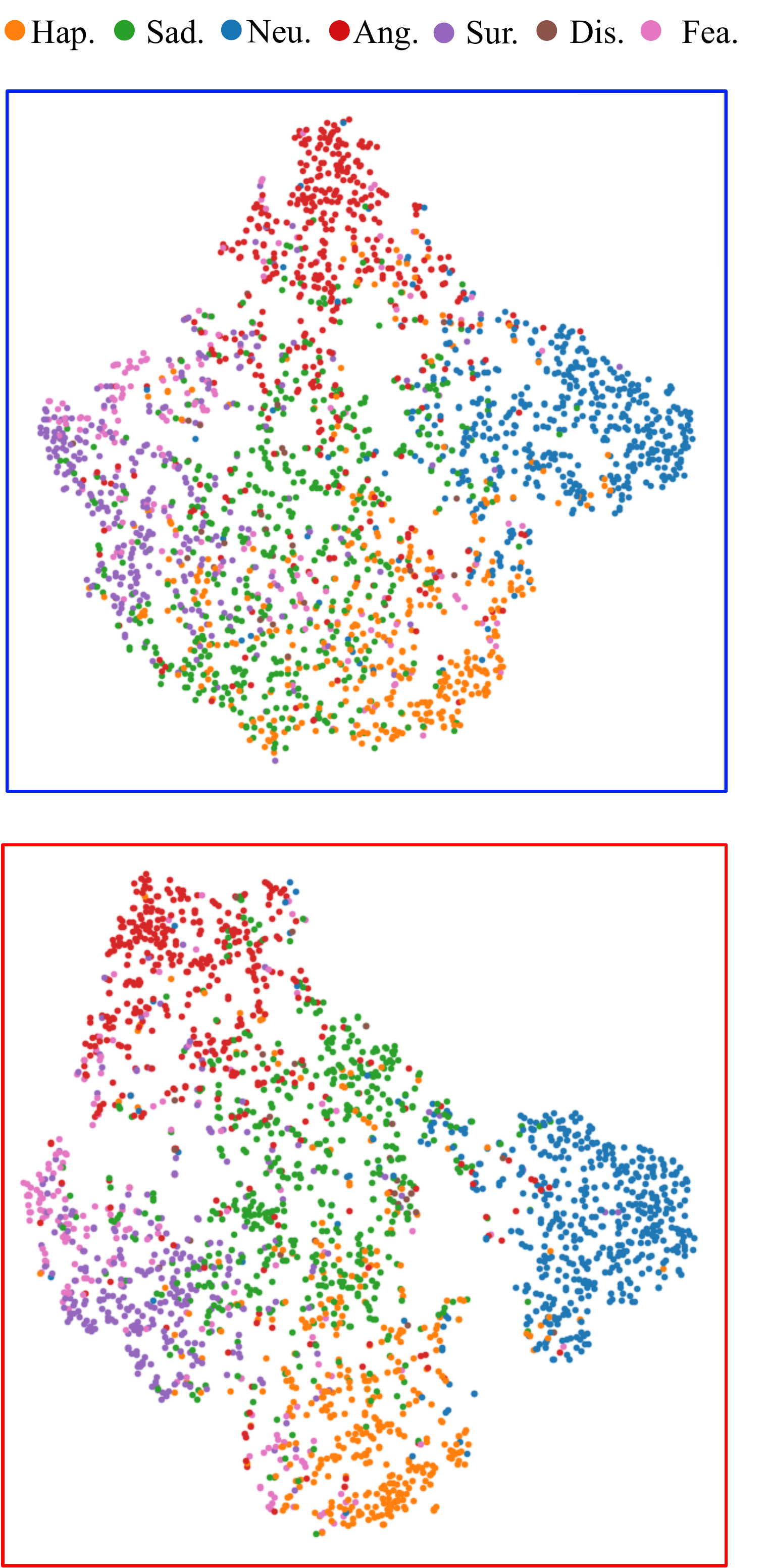} % Reduce the figure size so that it is slightly narrower than the column.
\caption{ Illustration of the learned feature distribution.}

\label{sne}
   % \vspace{-0.4cm}
\end{figure}

\subsection{Ablation Studies}

\subsubsection{Effectiveness of Each Module}
We conduct ablation studies to assess the contribution of each component in \name. As shown in Tab.~\ref{ablation}, the baseline without any module performs the weakest, highlighting the limitation of learning affective dynamics in a single entangled pathway. Introducing \MA~significantly improves both discrete and continuous metrics by enhancing low-rank stability and reducing noise-driven fluctuations. This supports our hypothesis that stabilizing long-term emotional bases provides a more coherent contextual field, which benefits both category decision making and continuous trajectory regression.

Adding \MB~brings further gains by decoupling sparse temporal perturbations and improving the discriminability of transient affect dynamics. Notably, the improvements after \MA$+$\MB suggest that stability and reactivity should not be optimized in a coupled manner: when transient fluctuations are routed and orthogonalized, the model becomes better at preserving short-lived discriminative cues without distorting the stable affective baseline.
The incorporation of \MC~provides additional improvements through cross-scale coordination, aligning global stability with local sensitivity. This indicates that hierarchical aggregation, while necessary for semantic abstraction, can introduce structural drift if stable and transient components are not explicitly reconciled across scales; \MC helps to maintain a consistent low-rank sparse organization throughout the network depth.

Finally, integrating \MD~yields the best overall performance by maintaining balanced optimization between smooth low-rank components and sensitive sparse dynamics. This result highlights that the benefit of hierarchical low-rank sparse modeling is not only architectural but also optimization-related: when gradient perturbations are adaptively regulated according to rank/sparsity-sensitive structure, training becomes more stable and generalization improves.
Overall, these observations confirm that each module fulfills a distinct yet complementary function, \MA stabilizes bases, \MB isolates transients, \MC enforces multi-scale coherence, and \MD improves structure-aware optimization, and their combination forms a coherent low-rank sparse equilibrium that supports both robust classification and accurate affect regression.

\subsection{Visualization.}

\subsubsection{Learned Feature Maps Visualization}
To further demonstrate the effectiveness of the proposed \name~(\textcolor{red}{red}) compared to the baseline (\textcolor{blue}{blue}) in modeling both discrete and continuous affective representations, we visualize the learned feature maps on DFEW and VEATIC. As shown in Fig.~\ref{vis}, \name~focuses more precisely on emotion-relevant facial areas such as the eyes and mouth, while effectively suppressing redundant background activations, indicating improved spatial compactness and semantic consistency brought by low-rank structural encoding.
Compared with the baseline, \name~exhibits cleaner attention distributions with reduced spurious responses on non-informative regions, which suggests that SEM-driven stabilization helps the model filter out high-frequency nuisance factors (\eg, illumination flicker, slight camera motion, background clutter) that frequently occur in real-world videos.
Moreover, \name~produces temporally smoother and spatially coherent activations across consecutive frames, highlighting the model’s capability to maintain stable affective dynamics while remaining responsive to short-term expression changes.
This behavior is consistent with the low-rank sparse principle: low-rank bases encourage coherent long-term patterns, while sparse transients allow localized bursts to be selectively emphasized rather than diffusely propagated.

In addition, the observed robustness under variations in illumination, pose, and expression intensity provides practical evidence that \name~can maintain stable attention even under capture variability, which is especially relevant to consumer electronics scenarios where cameras, viewpoints, and lighting conditions are highly diverse.
Overall, these results confirm that the low-rank and sparse aware mechanism enhances both spatial discriminability and temporal robustness in VAC, providing more reliable and interpretable affective evidence for downstream prediction.

\subsubsection{t-SNE Visualization}
We employ t-SNE \cite{SNE:v9:vandermaaten08a} to visualize the learned affective embeddings of the baseline (\textcolor{blue}{blue}) and \name~(\textcolor{red}{red}), as shown in Fig.~\ref{sne}.
The baseline produces scattered and overlapping clusters, indicating weak structural compactness and limited discriminability among affective states. This suggests that when stable trends and transient cues are entangled, the learned embedding space may encode mixed factors that blur the boundaries between affect categories and degrade regression consistency.

In contrast, \name~generates more distinguishable manifolds, where emotion categories exhibit tighter intra-class cohesion and more distinct inter-class boundaries.
The improved clustering pattern implies that \name~reduces representation redundancy and suppresses nuisance variations by organizing affective dynamics into a more structured low-rank (global context) and sparse (local bursts) composition.
As a result, samples from the same affect state become more compact even under large appearance diversity, while different affect states are better separated, reflecting stronger dynamic discriminability and structural consistency.
These observations qualitatively support our quantitative improvements on both WAR/UAR and RMSE, and further validate that hierarchical low-rank sparse modeling provides a more robust embedding geometry for video-based affective understanding.

%% file: sec/5_con.tex
 
\section{Conclusion}

We propose \name, a unified low-rank sparse representation learning framework for VAC.
By decomposing affective dynamics into low-rank stable structures and sparse transient variations, \name~achieves a structured representation of both persistent and fluctuating emotional patterns. 
Through hierarchical modules, \MA~for stable encoding, \MB~for dynamic disentanglement, \MC~for consistency integration, and \MD~for adaptive optimization, our framework establishes a principled bridge between robustness and plasticity.
Extensive experiments on discrete and continuous VAC datasets demonstrate its superior generalization under complex real world conditions.
In future, we will extend this paradigm toward multi-modal affective representation learning and efficient deployment for real time emotion analysis.